\documentclass[a4paper,fleqn,10pt,twocolumn]{SICE_ISCS}
\usepackage{amsmath,latexsym}
\usepackage[utf8]{inputenc}
\usepackage{graphicx}
\usepackage{caption, subcaption}
\usepackage{xcolor, soul}
\usepackage{multicol}
\usepackage{mathtools}
\usepackage{dsfont}

\title{Neural Continuous-Time Markov Models}

\author{Majerle Reeves${}^{\ast 1}$ and Harish~S. Bhat${}^{\ast 2}\dagger$}
\speaker{Harish~S. Bhat}

\affils{${}^{\ast}$Department of Applied Mathematics, University of California, Merced, 5200 N. Lake Rd, Merced, CA 95343 USA\\
${}^{1}$E-mail: mreeves3@ucmerced.edu\\
${}^{2}$E-mail: hbhat@ucmerced.edu
}
\abstract{
Continuous-time Markov chains are used to model stochastic systems where transitions can occur at irregular times, e.g., birth-death processes, chemical reaction networks, population dynamics, and gene regulatory networks. We develop a method to learn a continuous-time Markov chain's transition rate functions from fully observed time series. In contrast with existing methods, our method allows for transition rates to depend nonlinearly on both state variables and external covariates. The Gillespie algorithm is used to generate trajectories of stochastic systems where propensity functions (reaction rates) are known. Our method can be viewed as the inverse: given trajectories of a stochastic reaction network, we generate estimates of the propensity functions. While previous methods used linear or log-linear methods to link transition rates to covariates, we use neural networks, increasing the capacity and potential accuracy of learned models. In the chemical context, this enables the method to learn propensity functions from non-mass-action kinetics. We test our method with synthetic data generated from a variety of systems with known transition rates. We show that our method learns these transition rates with considerably more accuracy than log-linear methods, in terms of mean absolute error between ground truth and predicted transition rates.  We also demonstrate an application of our methods to open-loop control of a continuous-time Markov chain.}


\begin{document}

\maketitle

\section{Introduction}
Continuous-time Markov chains (CTMCs) are stochastic processes that model a variety of different systems, e.g., chemical reaction networks, population dynamics, gene regulatory networks, and infectious disease dynamics.  CTMC sample paths are piecewise-constant functions that take values in the state space; for a given sample path, each jump discontinuity corresponds to a time at which the system transitions from one state to the next.  On a discrete state space $\Omega$, a CTMC is completely characterized by its transition rates, an array $\Lambda = \{\lambda_{i,j}\}_{i,j \in \Omega}$ that describes the rate at which the system transitions from one state to another. Given an initial condition, which can be expressed as a distribution over the possible states, one can use the transition rates to generate sample paths of the system.   However, the inverse problem is not as straightforward.  Learning the transition rates from sample path data is an active field of research.

Our focus in this work is on a particular class of CTMCs that model the stochastic dynamics of  reaction networks \cite{Anderson2011}.  For this class of CTMCs, there has been a wealth of prior work on estimation and identification.  In the \emph{structure identification} problem, the goal is to estimate the reaction network itself.  Here we see approaches that focus on graphical modeling \cite{Cao2008}, sparse identification \cite{Braatz2014,klimovskaia2016sparse,Petzold2022}, and subnetwork reconstruction \cite{Yeung2015}.

Our work has more in common with the \emph{parameter estimation} problem, where all reactions and corresponding transition rates are known up to a finite set of parameters, which one seeks to estimate from data.  Various prior approaches can then be distinguished based on methodology and assumptions regarding the data.  In situations where one assumes access only to steady-state observations, we find both Bayesian \cite{gupta2019bayesian} and moment-based methods \cite{backenkohler2017moment}.  In the case where we assume access to inexact measurements of all chemical species, we see moment-based methods \cite{backenkohler2017moment}, variational Bayes methods \cite{Opper2010}, Gaussian adaptation methods \cite{Sbalzarini2012}, and expectation maximization methods, either for general systems \cite{horvath2008parameter}, or for the particular case of birth-death processes \cite{crawford2014estimation}.  There also exist filtering and Bayesian methods to estimate states and parameters from exact measurements of a subset of all chemical species in a given system \cite{Rathinam2021}, and adjoint methods for reaction-diffusion systems \cite{Tomlin2014}.

 In this paper, we model CTMC transition rates $\Lambda$ with \emph{functions} that we parameterize and estimate using neural networks.  We assume that all reactions are known, and that trajectories are fully observed.  Unlike prior parameter estimation studies we have seen, we assume nothing regarding the functional form of the transition rates (or, equivalently, propensity functions). The neural CTMC (N-CTMC) method we describe allows for transition rates with arbitrary kinetics and dependence on external covariates, such as temperature.
 
  Recent  work  has introduced covariate dependence via generalized linear models (GLM): for Markov chains in discrete/continuous-time,  transition probabilities/rates are modeled as a softmax/exponential function of a linear function of the covariates  \cite{Barratt2022,Awasthi2022}. In this paper, on tests with synthetic data, we show that the N-CTMC method can model kinetics that lie outside mass-action or Michaelis-Menten models. This goes beyond the capabilities of GLM models.
  
The likelihood function is often used to learn the transition rate matrix of a CTMC. The likelihood measures how probable a sample path is to occur given a transition rate matrix. To find CTMC transition rates that best fit one or more given sample paths, we solve for the transition rates that maximize the likelihood function.   For stochastic reaction networks with even a small number of reactions, the state space $\Omega$ can be large or infinite, rendering the likelihood intractable. Strategies to deal with the intractable likelihood include truncating the state space \cite{andreychenko2012approximate}, or using likelihood-free approaches such as approximate Bayesian computation \cite{owen2015scalable,toni2009approximate}. Truncating the state space introduces error, and does not allow for the full model to be learned.
   
 Instead of attempting to estimate a large or infinite-dimensional matrix $\Lambda$, N-CTMC uses the current state $i$ as one of the inputs to the neural network.  The neural network then outputs a vector of transition rates, one for each reaction.   In a setting with $s$ species of interest, every unique integer $s$-tuple $(A_1, \ldots, A_s)$ denotes a state. If each $A_i \in [0,N]$, the state space has maximal dimension $N^s$.  If we were to estimate each entry of $\Lambda$ individually, we would have to estimate $O(N^{2s })$ constant parameters.  In contrast, the number of parameters in the neural network model is independent of $N$.

 After deriving and explaining the method, we demonstrate its capabilities using tests with synthetic data.  In each test, we start with sample paths generated from a known reaction network.  We demonstrate that after training on these sample paths, N-CTMC succeeds in learning transition rates in close agreeement with the ground truth.  We explain and show how N-CTMC significantly reduces the computational cost of learning transition rates.  As we increase the sizes of our training sets, we show that the error of N-CTMC's estimates decreases. Finally, we show that N-CTMC significantly outperforms a GLM of a similar construction.

\section{Background}
\subsection{Stochastic Reaction Networks}
\label{sect:crn}
We first review an established framework for stochastic reaction networks. In general, a reaction network with $s$ species, $r$ reactions,  reaction rates $k$, and stoichiometric coefficients $\phi$ and $\psi$ can be written as:
\begin{equation}
\label{eqn:crn} \tag{$RN$}
\begin{split}
    \phi_{11} S_1 + \ldots + \phi_{s1} S_s &\xrightarrow{k_1} \psi_{11} S_1 + \ldots + \psi_{s1} S_s \\
    \phi_{12} S_1 + \ldots + \phi_{s2} S_s &\xrightarrow{k_2} \psi_{12} S_1 + \ldots + \psi_{s2} S_s \\
    & \! \quad \vdots \\
    \phi_{1r} S_1 + \ldots + \phi_{sr} S_s &\xrightarrow{k_r} \psi_{1r} S_1 + \ldots + \psi_{sr} S_s 
    \end{split}
\end{equation}
The state of the system at time $t$ is described by the state vector ${\mathbf{S}}(t)$, where $S_i(t)$ is a non-negative integer describing the count of species $i$ and ${\boldsymbol{\Delta} \mathbf{S}}_{\rho}$ denotes the change in the state after reaction $\rho$:
\[
\mathbf{S}(t) = \begin{bmatrix}
           S_{1}(t) \\
           S_{2}(t) \\
           \vdots \\
           S_{s}(t)
         \end{bmatrix}, \quad
{\boldsymbol{\Delta} \mathbf{S}}_{\rho} = \begin{bmatrix}
           \psi_{1\rho} - \phi_{1\rho} \\
           \psi_{2\rho} - \phi_{2\rho} \\
           \vdots \\
           \psi_{s\rho} - \phi_{s\rho}
         \end{bmatrix}
\]
Propensity functions describe how often reactions occur and are functions of the reaction rate and the current species count: $\alpha_i = g(k_i, {\mathbf{S}}(t))$. In mass action kinetics, $\alpha_i$ is often a polynomial function of ${\mathbf{S}}(t)$. The time until the next reaction $\tau$ is calculated using $\alpha = \sum_{i = 1}^r \alpha_i$, $\tau = (1/\alpha) \log (1/v_1)$ where $v_1 \sim U$, i.e., $v_1$ is sampled uniformly on the interval $[0,1]$. The next reaction is decided by sampling again, $v_2 \sim U$, and then determining for which $\rho \in \{1, 2, \ldots, r\}$ it is true that
\[
\frac{1}{\alpha} \sum_{i=1}^{\rho-1}\alpha_i \leq  v_2 < \frac{1}{\alpha} \sum_{i=1}^{\rho} \alpha_i.
\]
When $\rho=1$, we define the summation on the left to be $0$.  Once we have determined both $\tau$ and $\rho$, we update the state vector via  ${\mathbf{S}}(t + \tau) = {\mathbf{S}}(t) + {\boldsymbol{\Delta} \mathbf{S}}_{\rho}$.  


\subsection{Maximum Likelihood Estimation}
\label{sect:mle}
Before proceeding, we review the maximum likelihood estimate (MLE) of CTMC transition rates. Let $\Omega$ denote the state space for the CTMC.  Assume we have a sequence of state observations $O_i \in \Omega$ at times $t_i$:
\[
(O_0, t_0),(O_1, t_1), (O_2,t_2), ... (O_N, t_N).
\]
For states $l, m \in \Omega$ such that $l \neq m$, let $\lambda_{lm}$ denote the rate at which the CTMC transitions from state $l$ to state $m$.  The $\lambda_{lm}$ are the parameters to be estimated here.  Let $\lambda_{ll} = \sum_{m \in \Omega, m \neq l} \lambda_{lm}$.

 Given transition rates $\lambda$, the likelihood of observing the short sequence $(O_i, t_i), (O_{i+1}, t_{i+1})$ can be decomposed into two pieces. The first piece stems from the time $T_i = t_{i+1} - t_i$ spent in state $O_i$ before transitioning; this sojourn time has an $\mathrm{Exp}(\lambda_{O_i O_i})$ distribution.  The second piece is the probability of transitioning from state $O_i$ to another state $O_{i+1}$: this equals $\lambda_{O_i O_{i+1}} / \lambda_{O_i O_i}$.  Carrying this out across the entire observation sequence, we obtain the likelihood
\begin{equation}
\label{eqn:likelihood}
L(\lambda) = \prod_{i=0}^{N-1} 
 \lambda_{O_i O_i} e^{-\lambda_{O_i O_i} T_i}
 \frac{\lambda_{O_i O_{i+1}}}{\lambda_{O_i O_i}}.
\end{equation}
This likelihood can be rewritten in state space:
\begin{gather*}
{L}(\lambda)  =  \prod_{p \in \Omega} 
                   e ^{-\lambda_{pp} W_p}
                   \prod_{q \in \Omega, q \neq p}
                   \lambda_{pq}^{N_{pq}}\\
W_p   = \sum_{i=0}^{N-1} T_i I_{O_i = p}, \quad 
N_{pq} = \sum_{i=0}^{N-1} I_{O_i = p} I_{O_{i+1} = q}.
\end{gather*}
Here $W_p$ is the total time spent in state $p$, and $N_{pq}$ is the total number of transitions $p \rightarrow q$.  Because $\log$ is monotonic, we can maximize $L(\lambda)$ by maximizing
\[
 \log {L}(\lambda) = -\sum_{p \in \Omega} W_p \! \! \sum_{q \in \Omega, q \neq p} \! \! \lambda_{pq}  \, + \! \! \! \! \! \! \sum_{p,q \in \Omega, q \neq p} \! \! \! \! \! \! N_{pq} \log \lambda_{pq}.
\]
Differentiating with respect to $\lambda_{ab}$ and setting the derivative equal to zero, we obtain the MLE
\begin{equation}
\label{eqn:mle}
\begin{split}
\widehat{\lambda}_{ab} &= N_{ab} / W_a, \\
\widehat{\lambda}_{aa} &= \sum_{b \in \Omega, b \neq a} \widehat{\lambda}_{ab} = \left( {W_a}/ {\sum_{b \in \Omega, b \neq a} N_{ab}} \right)^{-1}.
\end{split}
\end{equation}
We see that $(\widehat{\lambda}_{aa})^{-1}$ is the mean time spent in state $a$.


\section{Methods and Metrics}
\label{sect:methodsandmetrics}
We now detail the neural CTMC (N-CTMC) method.
For a stochastic reaction network (\ref{eqn:crn}) as described in Section \ref{sect:crn}, assume we have complete observations:
${\mathcal{O}} = ({\mathbf{S}}(t_0),{\mathbf{C}}_0, t_0), \ldots ({\mathbf{S}}(t_N),{\mathbf{C}}_N, t_N)$.
Specifically, at time $t_i$, we have ${\mathbf{S}}(t_i) \in {\mathcal{R}}^s$, a vector of counts (of all species), and ${\mathbf{C}}_i \in {\mathcal{R}}^c$, the associated covariates.  Let ${\mathbf{x}}_i = [{\mathbf{S}}(t_i)^T, {\mathbf{C}}_i^T] \in {\mathcal{R}}^{s+c}$.

We look to model the propensity functions ${\boldsymbol{\alpha}}({\mathbf{x}}) \in {\mathcal{R}}^r$, where $r$ is the number of reactions in the reaction network (\ref{eqn:crn}). 
 In this formulation, we assume no knowledge of the reaction rates $k$.

We now formulate the likelihood function for the sequence of 3-tuples ${\mathcal{O}}$. In this likelihood function, we replace the constant transition rate matrix $\Lambda$ with a vector of propensity function ${\boldsymbol{\alpha}}({\mathbf{x}})$. To model ${\boldsymbol{\alpha}}({\mathbf{x}})$ we use a neural network, a universal function approximator, with parameters $\theta$.  The network takes as input ${\mathbf{x}}$ and outputs the propensity ${\boldsymbol{\alpha}}({\mathbf{x}}; \theta)$.  In contrast with Section \ref{sect:crn}, ${\boldsymbol{\alpha}}({\mathbf{x}}; \theta)$ depends not only on the  state ${\mathbf{S}}$ but also on the associated covariates ${\mathbf{C}}$.   Taking  the negative  $\log$ of (\ref{eqn:likelihood}) gives
\begin{equation}
\label{eqn:negloglik}
-\log L(\lambda) = \sum_{i=0}^{N-1} 
  \lambda_{O_i O_i} T_i - \log \lambda_{O_i O_{i+1}}.
\end{equation}
In the data ${\mathcal{O}}$, the transition $O_i \to O_{i+1}$ equates to the change in count vector ${\mathbf{S}}(t_i) \rightarrow {\mathbf{S}}(t_{i+1})$; let $\rho_i \in RN$ denote the reaction to which this change corresponds.  Then we can replace $\lambda_{O_i O_{i+1}}$ by the $\rho_i$-th element of the neural network's output, which we denote by $\alpha({\mathbf{x}}_i; \theta)_{\rho_i}$.  Similarly, we can replace $\lambda_{O_i O_i}$, the rate of leaving state $O_i$, by  $\sum_{j=1}^r \alpha({\mathbf{x}}_i; \theta)_j$, the sum of the propensity functions for each reaction in (\ref{eqn:crn}).  Rewritten in terms of   ${\mathbf{x}}$ and ${\boldsymbol{\alpha}}({\mathbf{x}}; \theta)$, the negative log likelihood (\ref{eqn:negloglik}) becomes
\begin{equation*}
-\log L(\theta) = \sum_{i=0}^{N-1} 
    \Big( T_i \sum_{j=1}^r \alpha({\mathbf{x}}_i; \theta)_j \Big)  
    - \log \alpha({\mathbf{x}}_i; \theta)_{\rho_i}. 
\end{equation*}
Let $L^{\rho}$ be the number of times reaction $\rho$ occurs.  Suppose we collect all row vectors ${\mathbf{x}}_i$ that correspond to states just before reaction $\rho$ occurs.  We stack these vertically into an $L^{\rho} \times (s+c)$ matrix $X^{\rho}$.  Similarly, let ${\mathbf{T}}^{\rho} \in {\mathcal{R}}^{ \, L^{\rho}}$ be a vector of time spent in state before reaction $\rho$ occurs. With this, we rewrite the above negative log likelihood as
\begin{multline}
\label{eqn:loss}
-\log L(\theta) = \sum_{\rho=1}^r \sum_{l = 1}^{L^{\rho}}
  {T}^{\rho}_{l} \Big( \sum_{j=1}^r \alpha({X}^{\rho}_{l} ; \theta)_{j} \Big) \\
  - \log \alpha({X}^{\rho}_{l} ; \theta)_{\rho}.
\end{multline}
\emph{To train the N-CTMC model, we apply gradient descent to minimize this negative log likelihood.}
Note that the internal sum over $L^{\rho}$ does not require a for loop in our implementation.  With neural network input vectorization, we can pass \emph{all} of $X^\rho$ into ${\boldsymbol{\alpha}}$ at once.  The $l$-th row of the output will be identical to the output we would have obtained had we passed in only $X^\rho_l$, i.e., $\alpha({X}^{\rho}; \theta)_{l j} = \alpha({X}^{\rho}_{l}; \theta)_{j}$.  Using this, the internal sum can be accomplished entirely with matrix algebra, which is highly optimized in modern neural network frameworks.  The only remaining loop corresponds to the sum over $\rho$ from $1$ to $r$; this should be contrasted with sums over $O(|\Omega|^2)$ elements that would have resulted from na\"{i}ve fusion of neural networks with the methods from Section \ref{sect:mle}.
Together with automatic differentiation, our formulation leads to more efficient computation of $\nabla_{\theta} \log L(\theta)$ and resulting minimization of the loss (\ref{eqn:loss}).


Note that the above N-CTMC approach learns a model ${\boldsymbol{\alpha}}({\mathbf{x}}; \theta)$ \emph{without} direct observations of ground truth propensity functions.  We observe the propensities only indirectly through their effect on the system, as evidenced in the data ${\mathcal{O}}$.  This varies from standard neural network training approaches where one starts with ground truth observations of a response variable and then minimizes the mean squared error beween predicted and true values of the response.

Here we briefly sketch how one might prove that neural networks approximations can converge to true propensity functions.  For brevity, let ${\boldsymbol{\alpha}} = {\boldsymbol{\alpha}}({\mathbf{x}})$.  Let  $\widehat{\boldsymbol{\alpha}}^{\theta}$ and $\widehat{\boldsymbol{\alpha}}^{N}$ denote propensity functions estimated by, respectively, a neural network and maximum likelihood estimation as in (\ref{eqn:mle}).  More specifically, we begin by binning the covariate space ${\mathcal{R}}^c$. For a given bin ${\mathcal{B}}$, using transitions ${\mathbf{S}}(t_i) \rightarrow {\mathbf{S}}(t_{i+1})$  from ${\mathcal{O}}$ such that  ${\mathbf{C}}_i \in {\mathcal{B}}$, we form MLE estimates as in (\ref{eqn:mle}).  Now given a state ${\mathbf{S}}$ and covariates ${\mathbf{C}}$, we find the bin corresponding to ${\mathbf{C}}$ and evaluate the MLE estimates (\ref{eqn:mle}) in that bin associated with all transitions of the form ${\mathbf{S}} \rightarrow {\mathbf{S}}'$.  Organizing these estimates by reaction number, we form $\widehat{\boldsymbol{\alpha}}^{N}$.  Let us assume that MLE consistency results for covariate-free CTMCs \cite{ranneby1978necessary} can be extended to include covariates, assuming conditions on the distribution of covariates ${\mathbf{C}}_i$ in the data ${\mathcal{O}}$ and on the ergodicity of the  CTMC, such that as $N \to \infty$, we will have $\| \widehat{\boldsymbol{\alpha}}^{N} -{\boldsymbol{\alpha}} \|_{\infty} \to 0$ almost surely (a.s.).  Next, treating $\widehat{\boldsymbol{\alpha}}^{N}$ as ground truth, we train a neural network with suitable activation functions, width $W$ and depth $D$; as $D \to \infty$, we have $\| \widehat{{\boldsymbol{\alpha}}}^{\theta} - \widehat{{\boldsymbol{\alpha}}}^N\|_{\infty} \to 0$ \cite{gripenberg2003approximation}.  Putting everything together, we obtain $\| {\boldsymbol{\alpha}}^{\theta} - {\boldsymbol{\alpha}}\|_{\infty} \to 0$ a.s. as both $N, D \to \infty$.




To evaluate the N-CTMC method, we will use several metrics: the mean absolute error (MAE) and mean squared error (MSE) across each unique state in ${\mathcal{O}}$, together with the weighted mean absolute error (W-MAE) and weighted mean squared error (W-MSE).  The W-MAE and W-MSE  examines the prevalence of each unique state in ${\mathcal{O}}$ and weights the contribution to the overall absolute error accordingly. This means that a state  that is rarely visited in ${\mathcal{O}}$ will have less effect on the W-MAE and W-MSE than on the MAE and MSE.


\section{Numerical Experiments}
We examine three different applications of the N-CTMC: a birth-death process, Lotka-Volterra population dynamics, and a temperature-dependent chemical reaction network.  Each application is chosen to test the N-CTMC model in a specific way.

In the birth-death process, we use a single covariate and let propensity functions depend solely on this covariate and not on the current population count. While this model lacks realism, it allows us to compare the N-CTMC to the counting-based MLE (\ref{eqn:mle}) for calculating rate parameters described in Section \ref{sect:mle}.
To test the N-CTMC's ability to capture arbitrary kinetics, we turn to a Lotka-Volterra population model.  Here there are no covariates, but the propensity functions do not follow the law of mass action. The final reaction network we consider tests the N-CTMC's ability to model arbitrary dependence on covariates.  Here the system of reactions follows the law of mass action, but the  propensities depend on temperature nonlinearly.

For each system, we generate trajectories using the Gillespie algorithm \cite{erban2007practical} and then use N-CTMC to learn propensity functions (equivalently, transition rates).  We model all systems with an increasing amount of training data to show empirical model convergence.

\begin{figure}[t]
\begin{center}
\includegraphics[width=8cm]{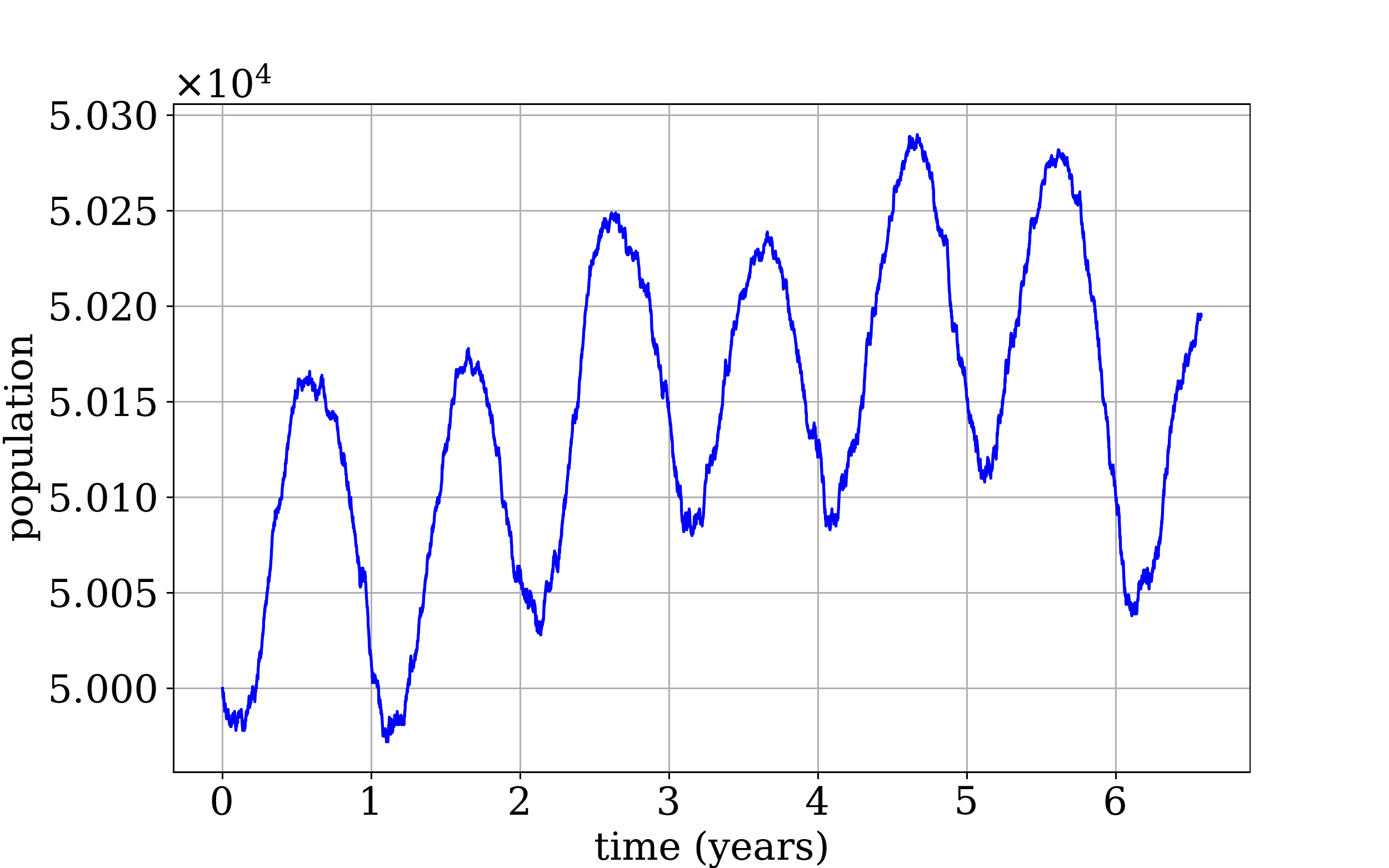}
\caption{\label{fig:BDP_traj} Sample trajectory from the birth-death process.}
\end{center}
\end{figure}

\subsection{Birth-Death Process}
We analyze a birth-death process where the birth and death rates do not depend on population size:
\vspace{-0.25cm}
\begin{multicols}{2}
\noindent Reactions:
   \begin{itemize}
                \item $\emptyset \rightarrow A$
                \item $A \rightarrow \emptyset$
   \end{itemize}
\columnbreak
Propensity Functions:
    \begin{itemize}
                \item $ \alpha_{1}(s) = 2.1 \textnormal{cos}(2 \pi s)$
                \item $ \alpha_{2}(s) = 2 \textnormal{sin}(2 \pi s)$
    \end{itemize}
\end{multicols}
\vspace{-0.25cm}
Let $t$ denote time in days.  Then $s = (t / 365.24) \textnormal{ mod } 1$, rounded to the tenths decimal place to discretize the covariate space to $s \in \{0.0, 0.1, 0.2, \ldots, 1.0\}$.  The birth and death rates $\alpha_1$ and $\alpha_2$ fluctuate annually, leading to sample trajectories such as that displayed in Figure \ref{fig:BDP_traj}.  We generate training trajectories of increasing length, $N = 5 \times 10^{\ell}$ for $\ell=3, 4, 5$, all with an initial population of $5 \times 10^4$, and train the N-CTMC model, yielding estimates $\widehat{{\boldsymbol{\alpha}}}^{\theta}(s)$.

While N-CTMC can handle continuous covariates, having a finite, discrete covariate space helps us compute transition rates using the counting-based MLE (\ref{eqn:mle}).  Specifically, for each distinct value $s'$ of the predictor, we find all transitions in the training set with predictor equal to $s'$.  Using those transitions, we compute the counting-based MLEs (\ref{eqn:mle}), resulting in estimates $\widehat{{\boldsymbol{\alpha}}}^N_j(s')$.

In Table \ref{tab:BD}, we compute in turn the MAE between the true ${\boldsymbol{\alpha}}$ and predicted transition rates $\widehat{{\boldsymbol{\alpha}}}^{\theta}$  and $\widehat{{\boldsymbol{\alpha}}}^{N}$ from the N-CTMC and counting approaches.  This yields, respectively, the errors labeled as N-MAE and C-MAE.  Carrying out the same comparison with MSE, we obtain the N-MSE and C-MSE errors.

Table \ref{tab:BD} shows convergence of both the N-CTMC  and counting-based estimated propensity functions to the true propensity functions, as the number of transitions increases. Table \ref{tab:BD} also shows that the error rates of the N-CTMC and counting-based methods are similar. This demonstrates that N-CTMC converges to the ground truth at a comparable rate to the  counting methods derived by analytically maximizing the likelihood. As the number of predictors increases, this calculation becomes more arduous.  Finally, Table \ref{tab:BD} gives evidence that, under appropriate hypotheses, a consistency result for covariate-dependent CTMCs should be possible.

\begin{table}[t]
\centering \small
\caption{Model performance for the birth-death process as the amount of data increases. Here we compare the N-CTMC (N) error with the counting-based MLE (C) calculation detailed in the main text.}
\begin{tabular}{l|l|l|l|l}
Transitions   & N-MAE                & C-MAE                & N-MSE                & C-MSE                 \\ \hline
5000          & $1.05 \cdot 10^{-1}$ & $1.07 \cdot 10^{-1}$ & $1.98 \cdot 10^{-2}$ & $1.89 \cdot 10^{-2}$   \\ 
50000         & $4.02 \cdot 10^{-2}$ & $3.97 \cdot 10^{-2}$ & $2.94 \cdot 10^{-3}$ & $2.95 \cdot 10^{-3}$   \\ 
500000        & $9.90 \cdot 10^{-3}$ & $9.76 \cdot 10^{-3}$ & $1.50 \cdot 10^{-4}$ & $1.60 \cdot 10^{-4}$   \\ 
\end{tabular}
\label{tab:BD}
\end{table}

\begin{figure}[b]
\begin{center}
\includegraphics[width=8cm,clip,trim=10 0 50 30]{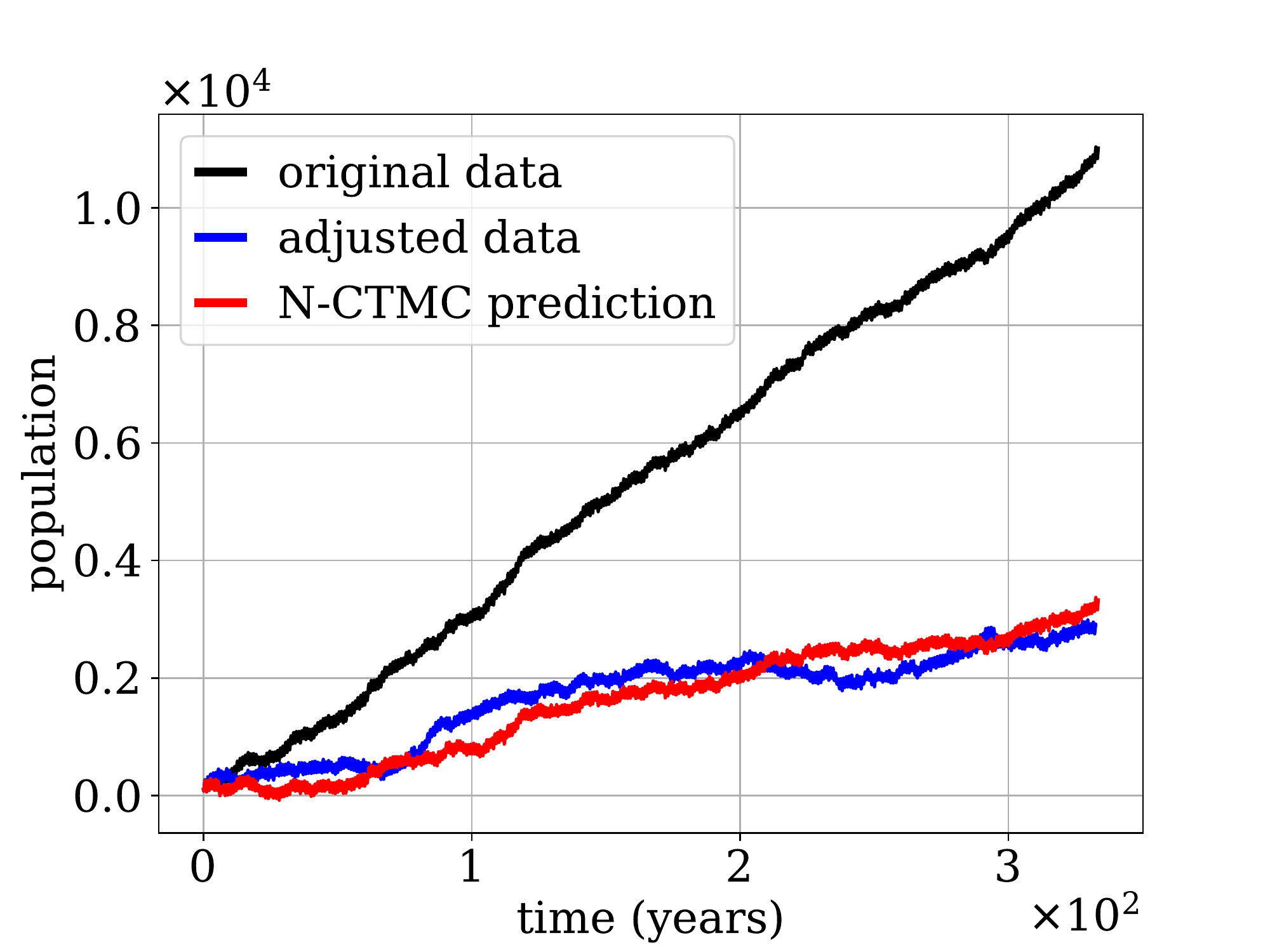}
\caption{We plot simulated population time series from the original birth-death process in black. We adjust the black curve, switching 1.5\% of the births to deaths (uniformly at random), resulting in the blue curve.  After training the N-CTMC on the blue curve, we generate a predicted time series, the red curve. The agreement between predicted and adjusted data is discussed in the main text.}
\label{fig:openloop}
\end{center}
\end{figure}

We now illustrate a potential open-loop control application of the N-CTMC method.  Assume we have a population of, for instance, a contagion/pest, that evolves stochastically according to the birth-death process described previously.  Suppose we simulate the system forward in time from an initial population of $100$ and find that it leads to too large of a population, the black curve in Figure 
 \ref{fig:openloop}.  What changes to the propensities are required to reduce population growth by a particular amount?  To answer: we 
process the data, switching 1.5\% of the births (uniformly at random) to deaths, resulting in the blue curve in Figure \ref{fig:openloop}.  \emph{We then train our N-CTMC model on the blue curve.}  With the estimated propensities, we then generate predictions starting from the same initial population of $100$---see the red curve in Figure \ref{fig:openloop}.  The close agreement indicates that N-CTMC succeeds in finding new propensity functions that, up to stochastic fluctuations, yield prescribed population dynamics.


\subsection{Population Dynamics Modeling}
Since neural networks are universal function approximators, N-CTMC should be able to model arbitrary transition rates. To test this, we consider a system that does not follow the law of mass action. We apply our method to a predator-prey system \cite{Enciso2021}, which models dynamics with prey (A) and two predators (B, C). This system includes contingencies for A, B, and C entering and leaving the system, breeding between species A, species B hunting species A, and species C hunting species B. 
\vspace{-0.25cm}
\begin{multicols}{2}
\noindent Reactions:
  \begin{itemize}
      \item $B \rightarrow \emptyset$
      \item $\emptyset \rightarrow B$
      \item $A \rightarrow \emptyset$
      \item $\emptyset \rightarrow A$
      \item $C \rightarrow \emptyset$
      \item $\emptyset \rightarrow C$
      \item $A \rightarrow 2A$
      \item $A + B \rightarrow 2B$
      \item $B + C \rightarrow 2C$
  \end{itemize}
\columnbreak
Propensity Functions:
    \begin{itemize}
        \item $ \alpha_{1} = B                        k_{1}     $
        \item $ \alpha_{2} =                          k_{2}     $
        \item $ \alpha_{3} = A                        k_{3} / N $
        \item $ \alpha_{4} =                          k_{4}     $
        \item $ \alpha_{5} = C                        k_{5}     $
        \item $ \alpha_{6} =                          k_{6}     $
        \item $ \alpha_{7} = A                        k_{7} / N $
        \item $ \alpha_{8} = A \sqrt{B}               k_{8}     $
        \item $ \alpha_{9} = \textnormal{log}(BC + 1) k_{9}     $
    \end{itemize}
\end{multicols} 
\vspace{-0.25cm}
Note that reaction 4 and 7 have the same outcome on the state of the system, so the neural network cannot distinguish between the two reactions. Hence these reactions will be modeled jointly: $\alpha_{4,7} = k_{4} + A  k_{7} / N$. Additionally, reactions 8 and 9 model hunting and do not follow the law of mass action.


\begin{figure}[t]
\begin{center}
\includegraphics[width=8cm]{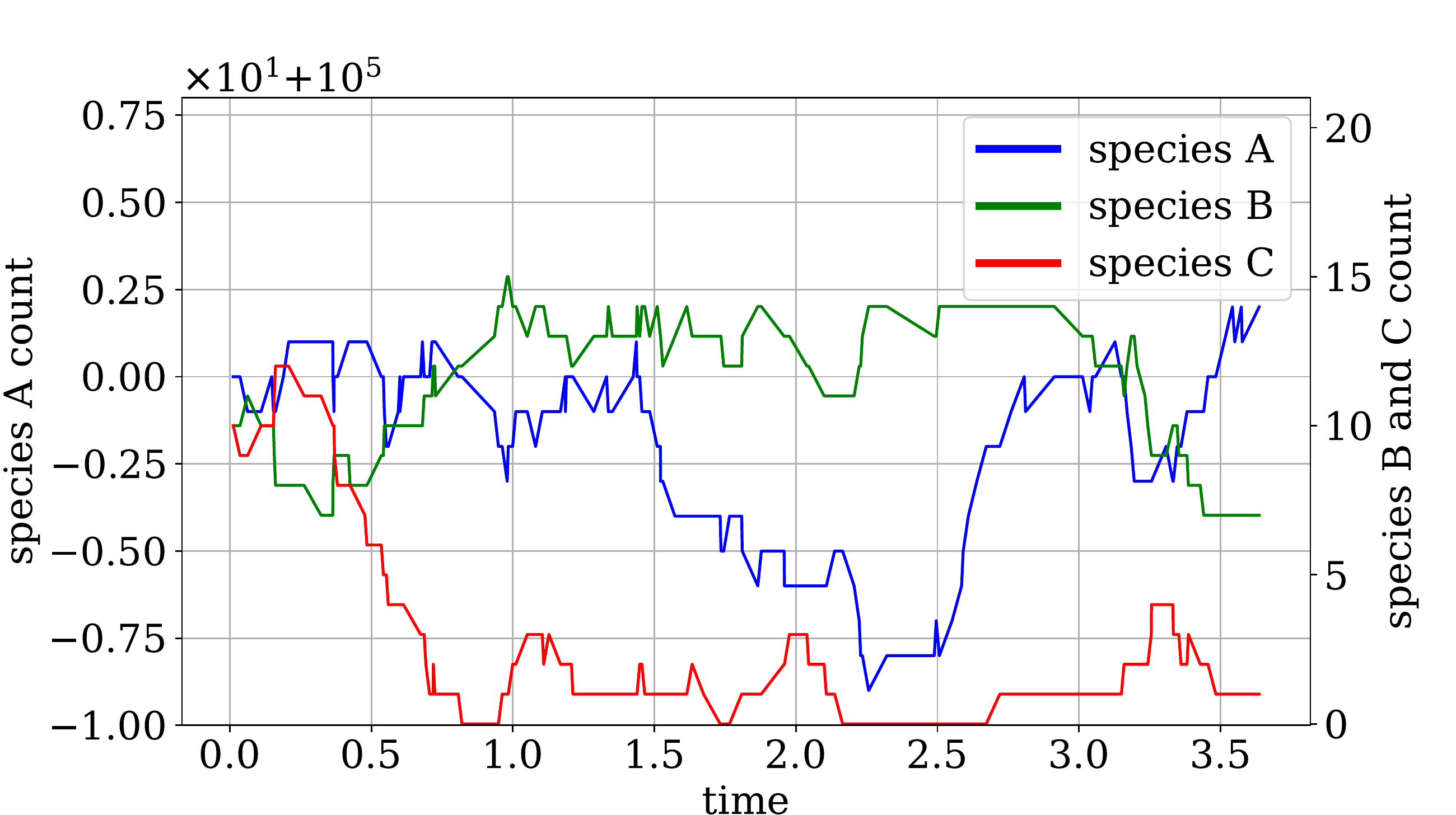}
\caption{\label{fig:LV_traj} One trajectory from the training data for the population dynamics system. We plot Species A on a separate axis from Species B and C due to the difference in scale.}
\end{center}
\end{figure}

We generate three training sets with, respectively, 100, 1000, and 10000 trajectories.  Each trajectory has length $N=150$, initial conditions  $(A(0),B(0),C(0)) = (10^5,10,10)$ and parameters ${\mathbf{k}} = [0.5, 1.7, 3.9, 4.6, 2.7, 1.9, 6.1, 2.4, 1.5]$. An example trajectory can be seen in Figure \ref{fig:LV_traj}. Using this data, we train both the N-CTMC model and, for the largest training set, a CTMC plus GLM model in which transition rates depend linearly on inputs. The N-CTMC models propensities via the following neural network:
\begin{align*}
     &z_1      = \sigma_1(x W_1 + b_1),  W_1 \in {{\mathcal{R}}}^{3 \times 128}, b_1 \in {{\mathcal{R}}}^{128}           \\
     &z_{i+1}  = \sigma_1(z_i W_{i+1}  + b_{i+1}), \quad 1 \leq i \leq 4   \\
     &W_{i+1} \in {{\mathcal{R}}}^{128 \times 128}, b_{i+1} \in {{\mathcal{R}}}^{128}, \quad 1 \leq i \leq 4   \\
     &\widehat{{\boldsymbol{\alpha}}}  = \sigma_2(z_5 W_{6} + b_{6}),    W_6 \in {{\mathcal{R}}}^{128 \times 9}, b_6 \in {{\mathcal{R}}}^{9}.   
\end{align*}
For $\sigma_1$, we use the scaled exponential linear unit (selu) activation function \cite{selu}. We set $\sigma_2(x)  = \log(1+e^x)$, the softplus function. We chose  softplus for $\sigma_2$ because the $\log \alpha$ in the loss (\ref{eqn:loss}) forces $\alpha > 0$. During model formulation, we tested several activation functions that guarantee positive output, and softplus performed the best.

\begin{figure*}[t]
     \centering
          \begin{subfigure}[b]{0.48\textwidth}
         \centering
         \includegraphics[trim={2cm 2cm 2cm 3cm},clip,width=\textwidth]{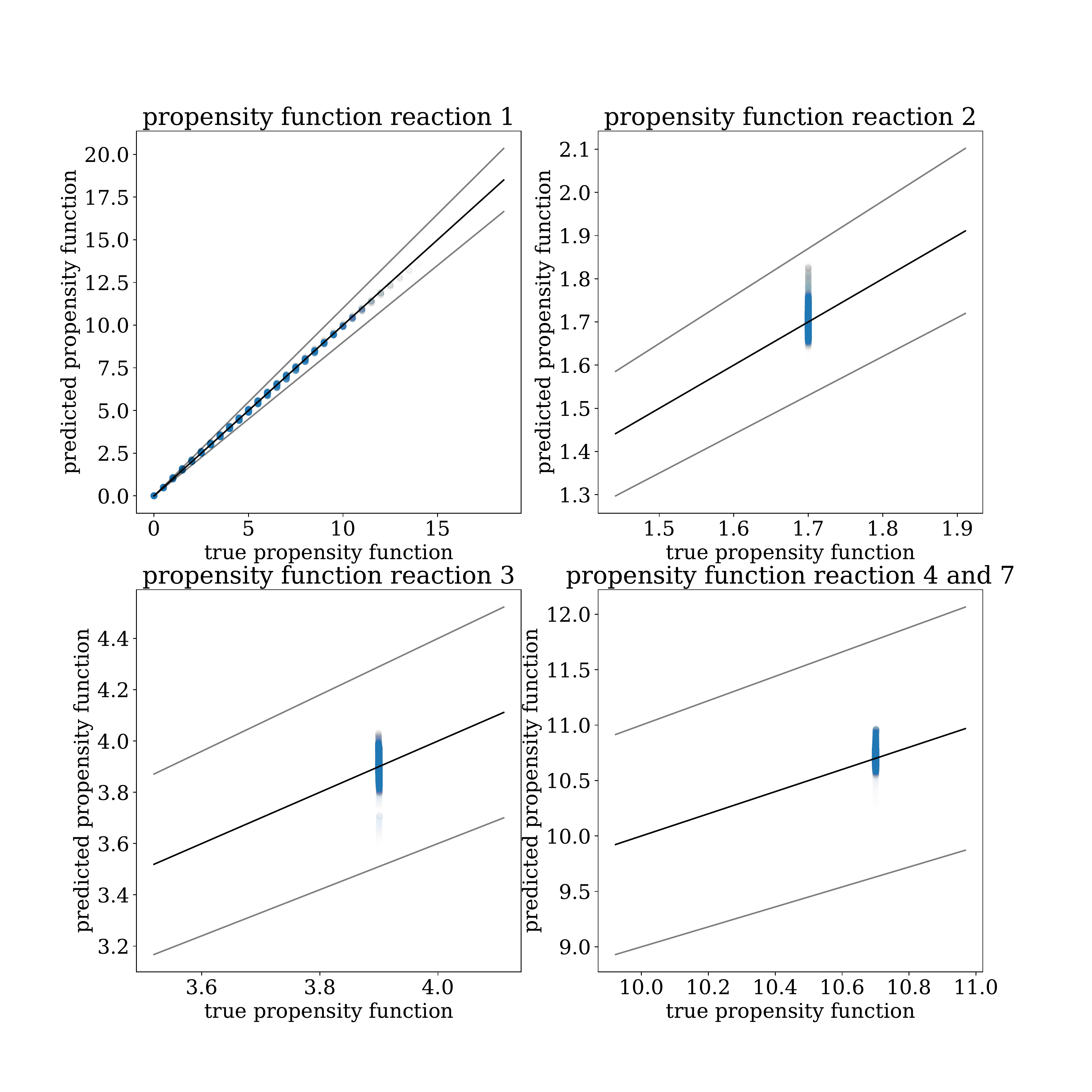}
         \caption{Predicted vs known reaction rates for reactions 1,2,3,4-7 where opacity denotes prevalence in the training set.}
         \label{fig:yequalsxopacity}
     \end{subfigure}
     \hfill
     \begin{subfigure}[b]{0.48\textwidth}
         \centering
         \includegraphics[trim={2cm 2cm 2cm 3cm},clip,width=\textwidth]{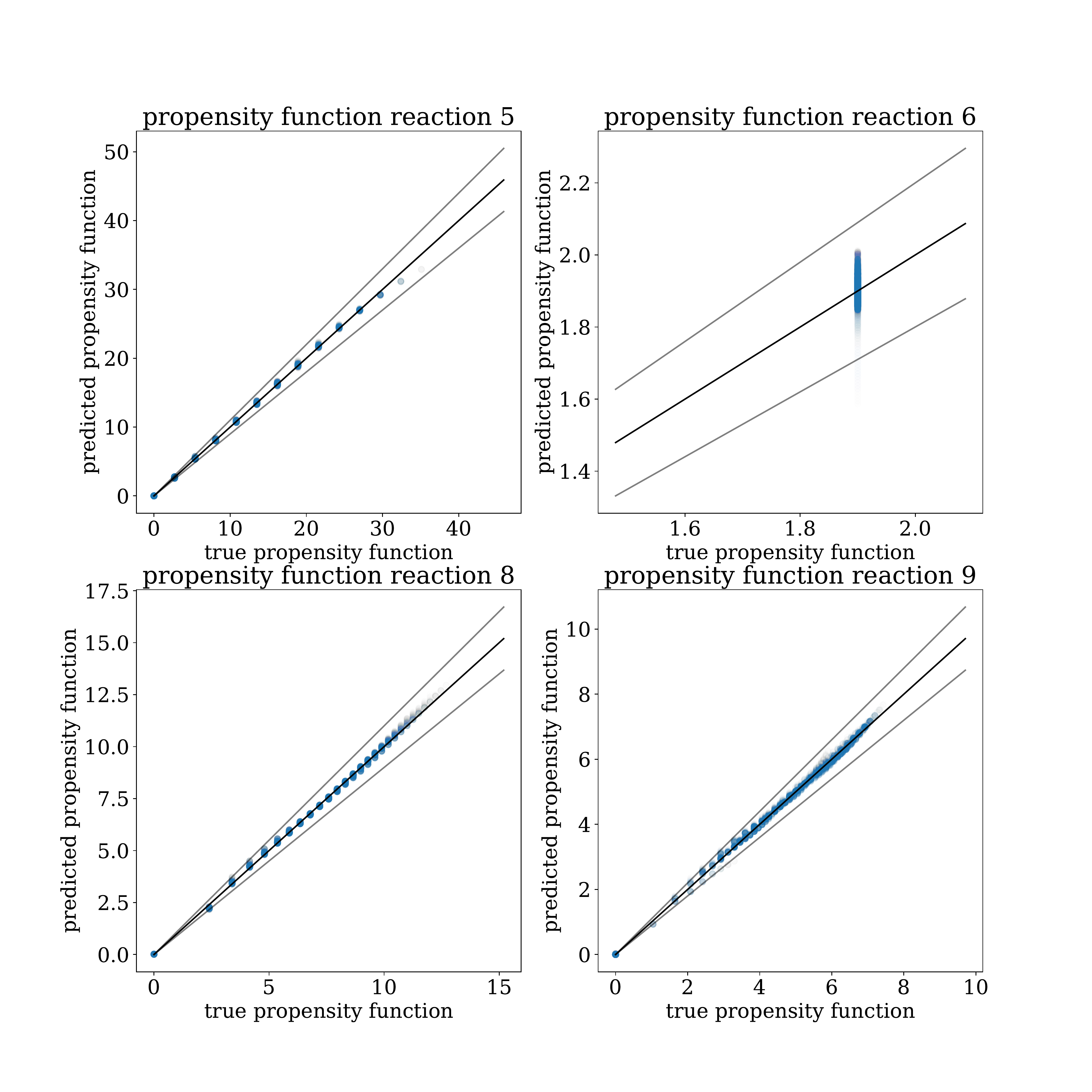}
         \caption{Predicted vs known reaction rates for reactions 5,6,8,9 where opacity denotes prevalence in the training set.}
         \label{fig:threesinxopacity}
     \end{subfigure}
     \caption{Comparison of reaction rates for the population dynamics modeling example. When plotting by prevalence in the training set in (a) and (b), the predicted values that are more prevalent in the training set achieve near-perfect accuracy (black line).}
     \label{fig:LVplot}
\end{figure*}

For each of the three training sets, we use
the four error metrics described at the end of Section \ref{sect:methodsandmetrics} to compare N-CTMC and GLM propensities against the true propensities.  As we see in Table \ref{tab:LV}, for the N-CTMC model, as the training set size increases, all error metrics decrease. This implies empirical convergence to the true model. From the weighted MAE and MSE metrics, we see that N-CTMC learns the true propensity very well for states that the trajectories visit frequently. The unweighted MAE and MSE metrics show that it is more difficult to learn propensity functions for rarely visited states. This can be easily visualized in Figure \ref{fig:LVplot} where predicted versus true propensity functions are plotted with intensity based on the prevalence of the state in the training data. When comparing the N-CTMC with the GLM, we see that the N-CTMC dramatically outperforms the GLM regardless of metric. Additionally, because the unweighted MAE and MSE are much larger for the GLM than for the N-CTMC, we conclude that the N-CTMC learns  propensities for rarely visited states much better than the GLM.

\begin{table}[t]
\centering \small
\caption{Model error for the population dynamics example.  The first three rows show the error of the N-CTMC model as we increase the number of trajectories in the training set. For the largest training set, we also show GLM model errors.  N-CTMC training is halted when $\Delta\text{loss} < 10^{-10}$. W denotes that the error metric is weighted. For the GLM model, training is halted when $\| \nabla \theta \| < 0.01$.} 
\begin{tabular}{l|l|l|l|l|l}
Trajectories & MAE   & W-MAE                 & MSE    & W-MSE                 & Epochs \\ \hline
100          & 0.304 & $2.45 \cdot 10^{-1}$  &  0.244 & $1.53 \cdot 10^{-1}$  & 12,900 \\ 
1000         & 0.157 & $1.01 \cdot 10^{-1}$  &  0.093 & $2.84 \cdot 10^{-2}$  & 33,500 \\ 
10000        & 0.099 & $3.82 \cdot 10^{-2}$  &  0.051 & $5.13 \cdot 10^{-3}$  & 89,800 \\ \hline
10000 (GLM)  & 1.028 & $7.06 \cdot 10^{-1}$  &  3.208 & $1.39 \cdot 10^{ 0}$  &  2,300 \\
\end{tabular}
\label{tab:LV}
\end{table}

For each of the three training sets, the number of unique states explored is 3767, 8380, and 13435, respectively. We see an increase in the state space size due to the stochastic nature of the trajectory generation: the more data generated, the more the state space is explored. Using counting-based likelihood-based approaches, we would need to estimate a transition rate for each distinct pair $(i,j)$ of states where both $i, j \in \Omega$ and $|\Omega| = 13435$.  Estimating each rate accurately would be highly inefficient and require a gargantuan amount of data.   The N-CTMC model has 67,721 parameters, as compared with the 107,480 non-zero elements of the transition rate matrix for the system with 13,435 states. By sharing propensity function parameters across $\Omega$, N-CTMC requires less data and less computational expense.


\subsection{Chemical Reaction Network Modeling}
\begin{figure}[b]
\begin{center}
\includegraphics[width=8cm]{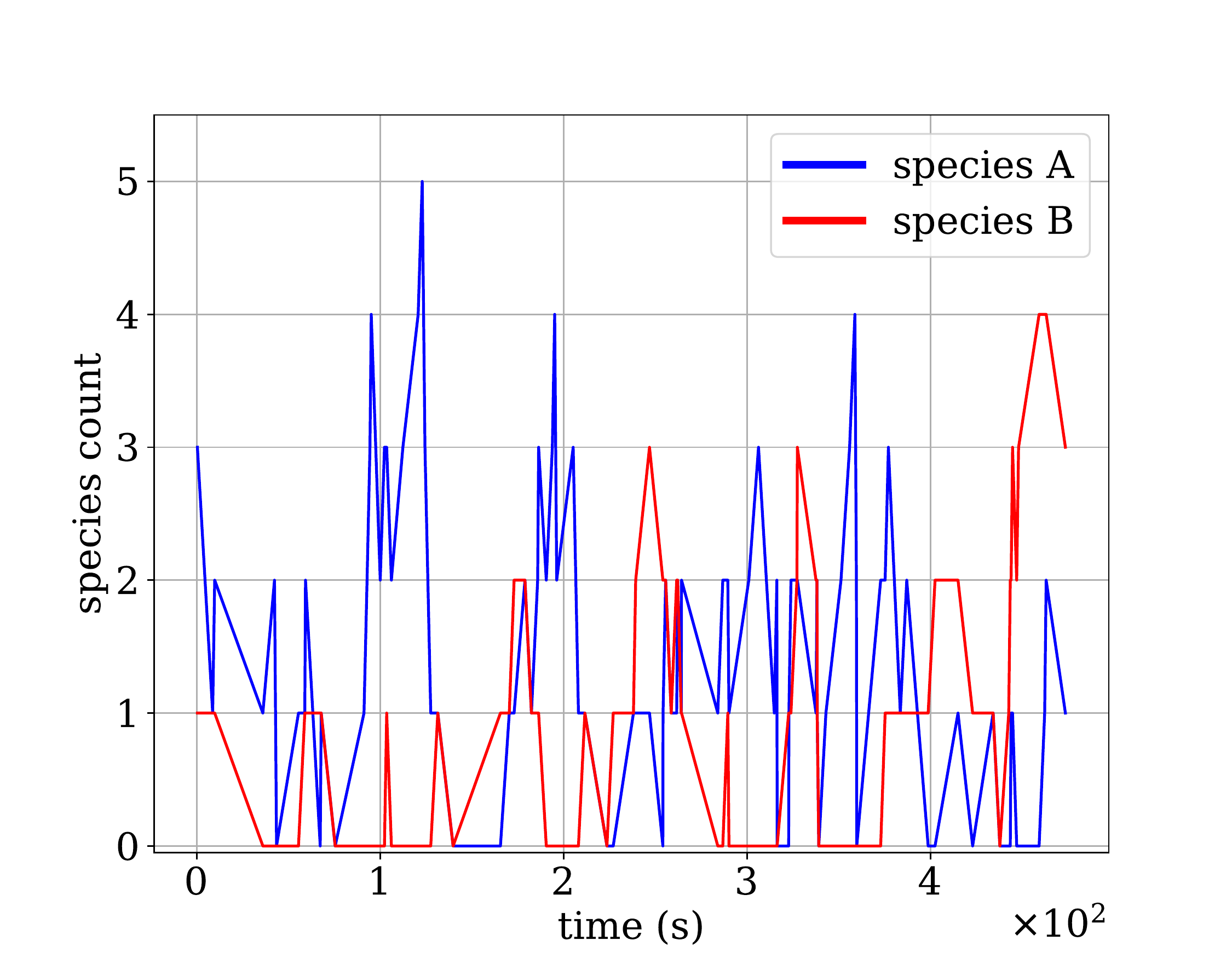}
\caption{\label{fig:CRN_traj} One trajectory (at the temperature 273 K) from the chemical reaction network training data.}
\end{center}
\end{figure}

In the third system, a modification of a system from \cite{erban2007practical}, we model the effect of temperature on a chemical reaction network. Reaction rates increase as a function of temperature according to $ \displaystyle k_j = {\mathcal{A}}_j e^{-E_j / (R T)}$, where ${\mathcal{A}}_j$ is the frequency factor, $E_j$ is the activation energy, $R$ is the ideal gas constant, and $T$ is the temperature in Kelvin. This network has two species, A and B:


\vspace{-0.25cm}
\begin{multicols}{2}
\noindent Reactions:
   \begin{itemize}
                \item $A + A \rightarrow \emptyset$
                \item $A + B \rightarrow \emptyset$
                \item $\emptyset \rightarrow A$
                \item $\emptyset \rightarrow B$
   \end{itemize}
\columnbreak
Propensity Functions:
    \begin{itemize}
                \item $ \alpha_{1} = A (A-1) k_{1} $
                \item $ \alpha_{2} = A B     k_{2} $
                \item $ \alpha_{3} =         k_{3} $
                \item $ \alpha_{4} =         k_{4} $
    \end{itemize}
\end{multicols}
\vspace{-0.25cm}
Note that ${\boldsymbol{\mathcal{A}}} = [630000, 770000, 5380000, 2240000]$ and ${\mathbf{E}} =  [39000, 36000, 40000, 40000]$.  For each fixed choice of $T$, we generate training sets with $20$, $60$, and $100$ trajectories, each with integer initial conditions $A[0], B[0]$ drawn uniformly $\in [0,4]$. This leads to $[100, 300, 500]$ total trajectories, as trajectories are generated at 5 different temperatures: $T = [271, 272, 273, 274, 275]$.  A sample trajectory at 273 K can be seen in Figure \ref{fig:CRN_traj}.
 Using this synthetic data, we train both a N-CTMC model and (using the largest training set only) a CTMC plus GLM model in which propensities depend linearly on inputs. 

In the N-CTMC, we use a convolutional neural network. We expand the inputs with a 96-unit dense layer, reshape into a $ 3 \times 32 $ matrix, and perform 1D convolutions across the rows with 10 output channels and a $ 1 \times 4 $ kernel according to $G[m] = (f*h)[m] = \sum_j h[j] f[m-j]$ where $f$ is the input and $h$ is the kernel. We then flatten the results and apply two linear layers with 32 nodes each, followed by a linear layer with 4 nodes. We use selu activation functions for all layers except the final layer where we use softplus to ensure positive output. Including a convolutional layer improved model performance over a dense neural network.  This model has 3,854 parameters, as compared to 5,820 parameters required in a na\"ive counting-based approach with 291 unique states and 4 reactions at 5 different temperatures. In this model, we implemented batch training, updating model parameters on each trajectory in turn, instead of on the whole data set at once. This improved model performance, possibly by allowing less frequently visited state spaces to have more impact on the overall likelihood gradients.

\begin{figure}[t]
\begin{center}
\includegraphics[trim={2cm 2cm 2cm 3cm},clip,width=8cm]{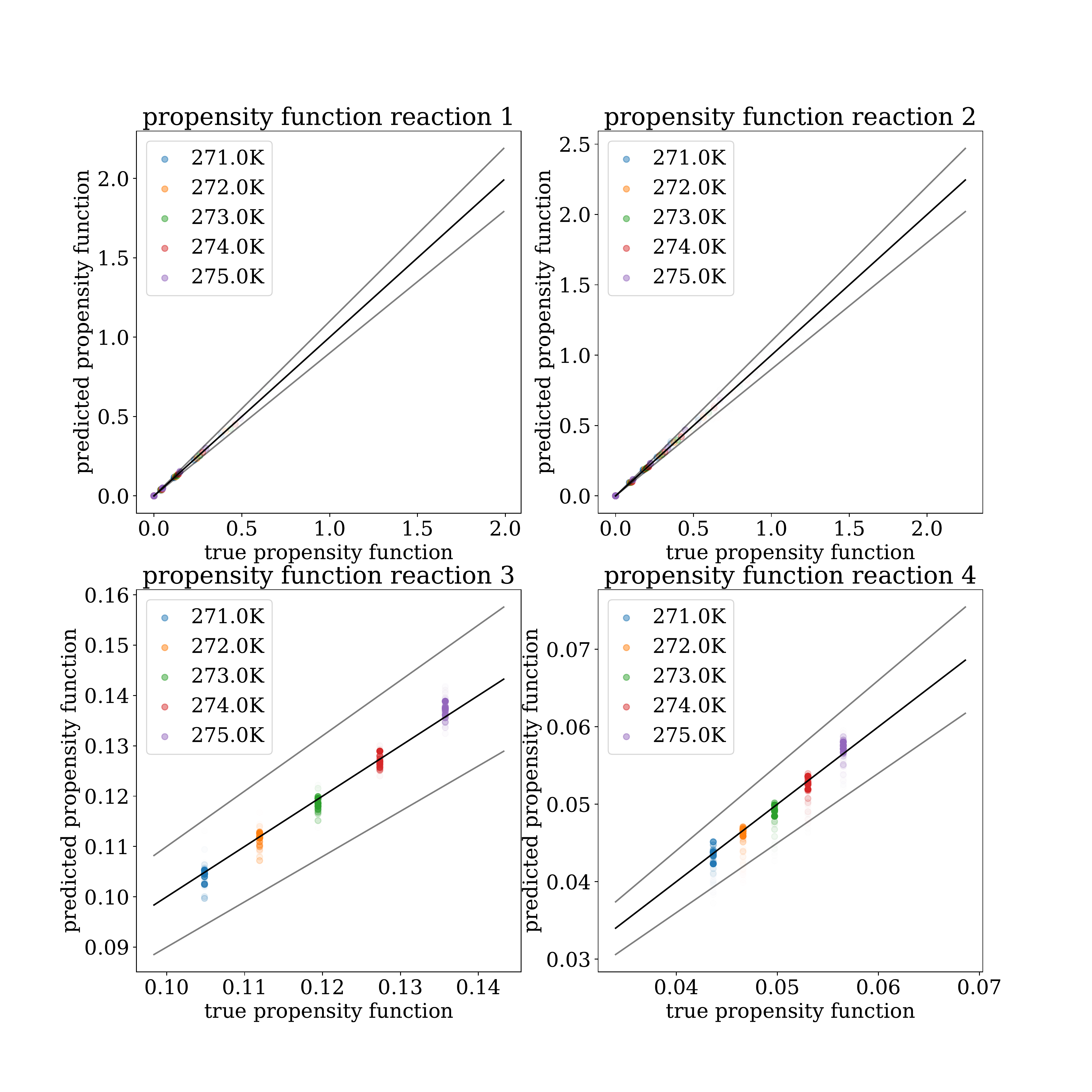}
\caption{\label{fig:CRNplot} Comparison of reaction rates for chemical reaction network modeling example. Opacity denotes prevalence of the corresponding states in the training set.  For more prevalent states, N-CTMC predicted propensity functions achieve near-perfect accuracy (black line).}
\end{center}
\end{figure}

In Table \ref{tab:CRN}, we quantify N-CTMC and GLM errors using the metrics described at the end of Section \ref{sect:methodsandmetrics}.  As we increase the number of trajectories, N-CTMC errors decrease, implying empirical convergence to ground truth propensity functions.   In Figure \ref{fig:CRNplot}, we plot N-CTMC predicted versus true transition rates, with varying intensities based on their prevalence in the training data.  Based on the W-MAE and W-MSE errors in Table \ref{tab:CRN} and the results in Figure \ref{fig:CRNplot}, our conclusions regarding the strong performance of N-CTMC on frequently visited states are the same as in the discussion of results for the population dynamics system above.  Across all metrics, the N-CTMC outperforms the GLM, and especially in learning rare state transition rates.

\begin{table}[t]
\centering \small
\caption{Model performance for the temperature-dependent chemical reaction network. The first three rows show the error of the N-CTMC model as we increase the number of trajectories in the training set. For the largest training set, we also show GLM model errors. Training is halted when $\Delta\text{loss}$ reaches a minimum. W denotes that the error metric is weighted. For the GLM model, training is halted when $\| \nabla \theta\| < 0.01$.} 
\begin{tabular}{l|l|l|l|l|l}
Trajectories & MAE   & W-MAE                  & MSE   & W-MSE                  & Epochs \\ \hline
100          & 0.024 & $ 3.13 \cdot 10^{-3}$  & 0.004 & $ 7.65 \cdot 10^{-5}$  &  7,700 \\ 
300          & 0.020 & $ 1.88 \cdot 10^{-3}$  & 0.003 & $ 2.61 \cdot 10^{-5}$  &  6,900 \\ 
500          & 0.015 & $ 1.45 \cdot 10^{-3}$  & 0.002 & $ 1.62 \cdot 10^{-5}$  &  4,000 \\ \hline
500 (GLM)    & 0.250 & $ 2.88 \cdot 10^{-2}$  & 1.345 & $ 5.80 \cdot 10^{-3}$  & 12,000 \\ 
\end{tabular}
\label{tab:CRN}
\end{table}

\section{Discussion}
We have assumed access to complete observations of trajectories together with knowledge of the number of reactions and associated stoichiometric coefficients.  Subject to these assumptions, we have demonstrated that the N-CTMC method learns consistent transition rates for stochastic reaction networks with both mass action and non-mass action kinetics, including for one system with more than 13,000 states. 

Unlike other methods where the likelihood is truncated, N-CTMC uses the entire state space in the likelihood calculation.  The N-CTMC estimated propensity function can be evaluated on any state in $\Omega$, including but not limited to states visited in the training set. 
We can only guarantee accurate estimates of propensity functions on states that have been visited sufficiently often in the training data. On rarely visited states, however, N-CTMC  outperformed a GLM approach in which propensity functions depend linearly on inputs (count vectors and/or covariates); such GLM approaches have been explored in prior work \cite{Barratt2022,Awasthi2022}.  The N-CTMC also models transition rates which vary as arbitrary functions of covariates, allowing for study of systems where propensity functions depend on external stimuli. 

In the current work, we focused on systems with exponentially distributed sojourn times, an unreasonable assumption for some systems \cite{Chiarugi2015}. Using the framework developed here, the negative log likelihood (NLL) can be easily modified to accommodate a semi-Markov process \cite{Asanjarani2022}.  For the birth-death example, we showed that N-CTMC can also be used to control---in an open-loop fashion---covariate-dependent CTMC systems.  We hope to explore both semi-Markov versions and control applications of N-CTMC in future work. 
 This will expand the class of problems to which we can apply our methods.



\smallskip

\setlength{\baselineskip}{-4pt} \setlength{\lineskip}{-4pt} \small \noindent \textbf{Acknowledgments.}
M. Reeves and H. S. Bhat acknowledge support from, respectively, NSF DGE-1633722 and NSF DMS-1723272. This research also benefited from computational resources that include the Pinnacles cluster at UC Merced (supported by NSF ACI-2019144) and  Nautilus, supported by the Pacific Research Platform (NSF ACI-1541349), CHASE-CI (NSF CNS-1730158), Towards a National Research Platform (NSF OAC-1826967), and UCOP.

\bibliographystyle{ieeetr}
\bibliography{main}

\end{document}